\documentclass[11pt]{article}

\usepackage[final]{latex/acl}
\usepackage{times}
\usepackage{latexsym}

\usepackage[T1]{fontenc}
\usepackage[utf8]{inputenc}

\usepackage{microtype}

\usepackage{inconsolata}

\usepackage{graphicx}
\usepackage{graphicx}      % 插入图片 (\includegraphics)
\usepackage{booktabs}      % 专业三线表 (\toprule, \midrule, \bottomrule)
\usepackage{amsmath}       % 数学公式 (\tilde, \setminus, \dots 等)
\usepackage{amssymb}       % 数学符号
\usepackage{multirow}
\usepackage{enumitem}
\title{When No Answer Is Correct:\\Diagnosing Absent Answer Detection for MLLMs in Video Understanding}

\author{
 \textbf{Yiheng Wang\textsuperscript{1}},
 \textbf{Yueqian Lin\textsuperscript{1}},
 \textbf{Lichen Zhu\textsuperscript{1}},
 \textbf{Yudong Liu\textsuperscript{1}},
 \textbf{Hai Li\textsuperscript{1}},
 \textbf{Yiran Chen\textsuperscript{1}}
\\
\\
 \textsuperscript{1}Duke University, Durham, NC, USA
 }

\begin{document}
\maketitle
\begin{abstract}
Multimodal large language models (MLLMs) have made substantial advancements in video understanding, yet the reliability of their responses remains underexplored.
This work presents a diagnostic study of absent answer detection for MLLMs in video understanding, where the correct answer is deliberately excluded from the candidate set and a reliable model is expected to recognize that no valid option exists.
We evaluate the absent answer detection behavior under three settings: multiple-choice questions augmented with a ``None of the Above'' option, open-ended generation with a detection instruction, and standard evaluation without any guidance.
Across a diverse set of models and benchmarks, we find that MLLMs overwhelmingly select plausible distractors rather than detecting the absent answer. This failure is more pronounced in temporal reasoning tasks and worsens with denser frame sampling.
We further explore chain-of-thought prompting as a mitigation strategy and find that while it substantially improves detection rates, performance remains unsatisfactory,  suggesting that prompting-based strategies alone are insufficient to fully address this limitation.
These findings expose a systematic failure in absent answer detection and highlight the need for explicit detection mechanisms in multimodal systems.
\end{abstract}
\section{Introduction}
\begin{figure}[t]
\centering
\includegraphics[width=1\linewidth]{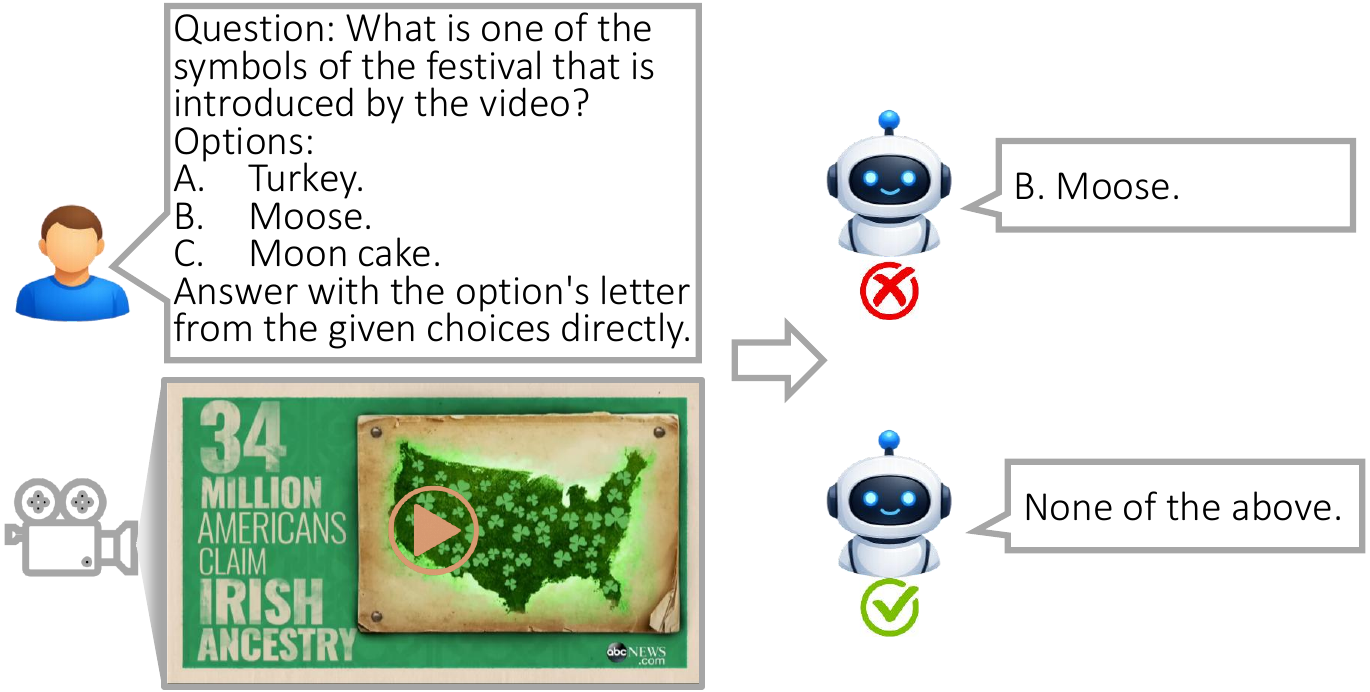}
\caption{An example of absent answer detection. Reliable MLLMs should recognize that no valid option exists rather than selecting a distractor.}
\vspace{-15pt}
\label{fig:motivation}
\end{figure}

Multimodal large language models (MLLMs) \cite{qwen2vl, qwen25vl, qwen25omni, internvl3, internvl35, gemma3, mimo-vl} have achieved remarkable progress in video understanding, demonstrating strong performance across a wide range of benchmarks~\cite{videomme, egoschema, longvideobench}. However, these benchmarks share a common assumption: the candidate set always contains the correct answer. Under this assumption, it remains unclear whether models truly understand the video content or merely select the most plausible option from the candidates. To probe this, we adopt a diagnostic approach by deliberately removing the correct answer from the candidate set. If a model genuinely understands the video and the question, it should be able to recognize that no valid option exists, rather than defaulting to the most plausible distractor. We term this diagnostic setting \textit{absent answer detection}. An example is depicted in Fig.~\ref{fig:motivation}.
% Multimodal large language models (MLLMs) \cite{qwen2vl, qwen25vl, qwen25omni, internvl3, internvl35, gemma3, mimo-vl} have achieved remarkable progress in video understanding, demonstrating strong performance across a wide range of benchmarks~\cite{videomme,egoschema,longvideobench}. 
% A reliable MLLM should not only provide correct answers when possible, but also recognize when a question cannot be reliably answered given the available options.

%: given a video, a question, and a candidate set from which the correct answer has been removed, a reliable model should recognize that no valid option exists. 
While absent answer detection has been explored in text-only and image-based settings~\cite{beyond_answers, not_an_option, UnsolvableProblemDetection}, video understanding, where temporal reasoning and multi-frame integration introduce additional complexity, remains unexamined. To study this problem, we remove the ground-truth answer from the candidate set while keeping the video and question unchanged, and evaluate models under three complementary settings: 
(1) multiple-choice questions augmented with a ``None of the Above'' option;
%, which provides an explicit detection cue; 
(2) open-ended generation with explicit instructions permitting the model to indicate that no option is correct;
%, which directly probes detection capability; 
and (3) unprompted evaluation under the standard protocol, which tests whether models can spontaneously detect the absence of a correct answer without any guidance. 
%Across a diverse set of models and benchmarks, we identify the following consistent findings:
Across a diverse set of models and benchmarks, we diagnose consistent failure patterns and explore chain-of-thought prompting as a potential mitigation. Our key findings are summarized as:
\begin{itemize}[itemsep=2pt, parsep=0pt, topsep=2pt]
    \item MLLMs overwhelmingly select plausible distractors rather than detecting the absent answer, and this limited detection ability is largely contingent on explicit external cues.
    \item Detection rates drop substantially on temporal tasks, where distractors are substantially more semantically similar to the removed ground-truth answer.
    %Detection rates drop substantially on temporal reasoning tasks, where temporally adjacent events serve as highly plausible distractors that obscure the absence of the correct answer.
    \item Denser frame sampling paradoxically worsens detection, suggesting that richer visual input strengthens candidate matching but suppresses critical evaluation of the option set.
    \item Chain-of-thought prompting substantially improves detection by encouraging per-option verification against the video content, yet detection rates remain unsatisfactory.
    %, suggesting that addressing this limitation will require interventions beyond inference-time strategies.
\end{itemize}
\section{Related Work}
\noindent \textbf{Video understanding with MLLMs.} Multimodal large language models~\cite{qwen2vl, qwen25vl, internvl3, internvl35,gemma3} have achieved strong performance across diverse video understanding benchmarks~\cite{videomme, longvideobench, mvbench, videommmu}. Recent studies have further explored efficient long-form video understanding through visual token pruning and memory, aiming to preserve informative temporal evidence under constrained context budgets~\citep{kvtp,hippomm}.
However, growing evidence suggests that high accuracy may not reflect faithful understanding. \citet{mirage} show that multimodal models can achieve competitive benchmark scores through language priors alone, without any visual input. In the video domain, \citet{virtuebench} and \citet{vllm_refuse} further probe model reliability by examining whether models can refuse to answer when key frames are missing or when questions exceed the video's informational scope. These efforts share a common focus on scenarios where the question itself is unanswerable. We instead introduce absent answer detection to video understanding, investigating whether models genuinely comprehend video content or merely select the most plausible candidate.
% \noindent \textbf{Video understanding with MLLMs} A growing body of work evaluates MLLMs in video understanding through MCQ-based benchmarks~\cite{videomme,longvideobench,mvbench,videommmu}. 
% Recent efforts acknowledge that MCQ accuracy alone does not reflect faithful understanding~\cite{videomme_v2, herbench}, yet these benchmarks still assume that the candidate set always contains a valid answer, leaving model behavior under absent answer conditions unexamined.

\noindent\textbf{Absent answer detection.}
Absent answer detection has been studied in text-only settings,
where \citet{beyond_answers} and \citet{nota_less_right} show that
replacing the correct option with ``None of the above'' causes
a substantial accuracy drop, and \citet{not_an_option} further
demonstrate that instruction tuning suppresses rather than promotes
the detection of invalid option sets. In the image domain,
\citet{UnsolvableProblemDetection} formalize incompatible answer
set detection for image MLLMs and show that models overwhelmingly
select a distractor instead of recognizing the absent answer.
UNK-VQA further evaluates the abstention ability of multimodal models on unanswerable visual questions~\cite{unkvqa}. Another study~\cite{zheng2024large} show that LLMs are sensitive to option positions in multiple-choice questions due to selection bias.
However, these studies are confined to text-only or image-based
settings. We extend the investigation to the video modality and uncover failure patterns unique to
video understanding.
\section{Experimental Setup}
\subsection{Evaluation Settings}
\textbf{Baseline setting.}
Given a video $V$, a question $q$, and the original MCQ candidate set $C = \{c_1, \dots, c_k\}$ containing the ground-truth answer $c^{*}$, we evaluate the model under the standard protocol to obtain baseline accuracy (\textbf{ACC}).

\noindent \textbf{Intervened setting.} To study absent answer detection, we construct an intervened candidate set $\tilde{C} = C \setminus \{c^{*}\}$ by removing the ground-truth answer while keeping the video and question unchanged. We then evaluate models under three conditions. The examples are provided in Appendix~\ref{appendix: prompt}.

\noindent \textbf{(1) Multi-choice detection.} We add a ``None of the above'' (NOTA) option to $\tilde{C}$, yielding a candidate set of size $k$ where NOTA serves as the correct answer and acts as an explicit detection cue. We report Multi-Choice Detection Rate (\textbf{MCDR}), the proportion of instances in which the model selects the NOTA option.
Because multiple-choice models may exhibit position bias, we additionally evaluate MCDR when placing NOTA at each candidate position while randomizing the remaining options. Appendix~\ref{app:position} shows that our conclusion is robust to the NOTA position.

\noindent \textbf{(2) Open-ended detection.} We prompt the model with the question and intervened candidate set $\tilde{C}$ alongside an instruction stating that the model may respond none if it determines that no option is correct. This setting removes selection-side biases and directly probes whether the model can detect the absent answer through free-form generation. We report Open-Ended Detection Rate (\textbf{OEDR}), the proportion of instances in which the model explicitly indicates that no option is correct.

\noindent \textbf{(3) Unprompted detection.} We present the model with the intervened candidate set $\tilde{C}$ under the standard evaluation protocol, i.e., the model is simply asked to select an answer from the given options, without any detection cue or instruction. We report Unprompted Detection Rate (\textbf{UDR}), the proportion of instances in which the model spontaneously detects the absent answer.

\subsection{Implementation Details}
% \textbf{Benchmarks.}
% We evaluate MLLMs on two widely used video understanding benchmarks: VideoMME~\cite{videomme} and EgoSchema validation set~\cite{egoschema}. 
% VideoMME covers diverse video understanding tasks with multiple-choice questions spanning different video durations. 
% EgoSchema focuses on egocentric video understanding, requiring models to reason about human activities and daily interactions from a first-person perspective.

% \noindent\textbf{Implementation Details.}
We evaluate representative MLLMs spanning proprietary models~\cite{gemini25,gpt4o} and open-source models~\cite{gemma3,qwen2vl, qwen25vl, qwen3vl, qwen25omni, qwen3omni,internvl3,internvl35,mimo-vl} on VideoMME~\cite{videomme} and EgoSchema validation set~\cite{egoschema}. All models use a default frame budget of $64$ frames unless otherwise stated, with greedy decoding for reproducibility.
Experiments are conducted on NVIDIA L40S GPUs by default, except for Qwen3-Omni-30B-A3B~\cite{qwen35omni}, which is evaluated on NVIDIA H100 GPU.
For models whose responses are short and reduce to an option letter, we extract predictions using rule-based regular-expression matching. For CoT-prompted models and Gemini-2.5-Flash, whose responses can be long and free-form, we use Qwen3-1.7B~\cite{qwen3} as an answer extractor to identify the selected option or explicit absent-answer behavior from the complete response. We manually validate the extraction procedure on a random subset of raw responses.
%We use the official inference settings for each model without additional fine-tuning, and keep decoding configurations consistent across benchmarks to ensure fair comparison.

%%%%%%%%%%%%%%%%%%%%%%%%%%%%%%%
%%%%%%%%% Result
%%%%%%%%%%%%%%%%%%%%%%%%%%%%%%%
\section{Experimental Results}

\begin{table}[t]
\centering
\small
\setlength{\tabcolsep}{4pt}
\begin{tabular}{l c cccc}
\toprule
Model & Scale & ACC & MCDR & OEDR & UDR\\
\midrule
Gemini-2.5-Flash & - & 68.9 & 33.9 & 43.6 & 2.4\\
\midrule
Gemma3 & 4B & 52.1 & 7.9 & 6.0 & 0.0 \\
Gemma3$^\dagger$ & 12B & 60.5 & 25.4 & 11.8 & 0.1 \\
Qwen2-VL & 7B & 61.3 & 13.1 & 1.2 & 0.0\\
Qwen2.5-VL & 7B & 62.8  & 11.7 & 11.9 & 0.0\\
Qwen2.5-Omni & 7B & 63.7 & 19.9 & 61.7 & 0.0\\
Qwen3-VL & 8B & 67.0 & 17.4 & 16.2 & 0.7\\
Qwen3-Omni & 30B & 68.1 & 23.0 & 9.3 & 1.9 \\
Mimo-VL & 7B & 64.3 & 40.9 & 8.8 & 0.0\\
InternVL3 & 8B & 66.4 & 11.9 & 1.1 & 0.0  \\
InternVL3.5 & 8B & 65.3& 13.6 & 6.5 & 0.0 \\
\bottomrule
\end{tabular}
\vspace{-5pt}
\caption{Absent answer detection performance on Video-MME. $\dagger$ denotes that the model is evaluated with 32 frames of input. }
\vspace{-5pt}
\label{tab:videomme}
\end{table}

\begin{table}[t]
\centering
\small
\setlength{\tabcolsep}{4pt}
\begin{tabular}{l c cccc}
\toprule
Model & Scale & ACC & MCDR & OEDR & UDR \\
\midrule
Gemini-2.5-Flash & - & 68.9 & 27.5 & 34.6 & 0.4\\
Gemini-2.5-Pro   & - & 72.0 & 60.6 & 52.6 & 1.4 \\
GPT-4o           & - & 65.6 & 31.0 & 28.0 & 1.0 \\
\midrule
% Gemma3 & 4B &  \\
% Gemma3^\ddagger & 12B & \\
Qwen2-VL & 7B & 66.4 & 3.2 & 1.6 & 0.0 \\
Qwen2.5-VL & 7B & 66.0 & 8.8 & 9.0 & 0.0 \\
% Qwen2.5-Omni & 7B \\
Qwen3-VL & 8B & 74.6 & 21.0 & 23.2 & 1.6\\
Qwen3-Omni & 30B & 66.4 & 26.6 & 2.8 & 0.2 \\
Mimo-VL & 7B & 68.0 & 40.6 & 17.4 & 0.0 \\
InternVL3 & 8B & 79.2 & 9.6 & 2.2 & 0.0 \\
InternVL3.5 & 8B & 74.8 & 11.0 & 5.8 & 0.0 \\

\bottomrule
\end{tabular}
\vspace{-5pt}
\caption{Absent answer detection performance on EgoSchema.}
\vspace{-10pt}
\label{tab:egoschema}
\end{table}

\subsection{Detection Behavior Analysis}
Table~\ref{tab:videomme} and Table~\ref{tab:egoschema} report the absent answer detection results on VideoMME and EgoSchema.

% \noindent\textbf{Multi-choice detection.} Despite strong baseline accuracy, all models exhibit markedly low MCDR, raising the question of whether models genuinely know the correct answer or merely select the most plausible option from the candidate set. To investigate this, we examine the confidence distributions under the multi-choice detection setting (Appendix~\ref{appendix: confidence distribution}). We find that models maintain near-baseline confidence levels even after the ground-truth answer is removed, exhibiting pronounced overconfidence across the majority of instances. Moreover, the probability mass assigned to the remaining distractors is substantially higher than that assigned to the NOTA option. These findings suggest that models predominantly default to selecting the most plausible candidate rather than truly knowing the correct answer and detecting its absence. Nonetheless, stratifying by baseline correctness (Appendix~\ref{appendix: statistical analysis}) reveals that models possess a weak sensitivity to answer absence, though it is far from sufficient to drive reliable detection behavior: models are significantly more likely to select NOTA on questions answered correctly at baseline (OR: 2.7--5.5, $p < 10^{-10}$), yet the small effect sizes ($\varphi = 0.13$--$0.26$) indicate that this signal rarely translates into consistent detection in practice.

\noindent\textbf{Multi-choice detection.} Despite strong baseline accuracy, all models exhibit markedly low MCDR. We examine the confidence distributions under the multi-choice detection setting (Appendix~\ref{appendix: confidence distribution}) and find that models maintain near-baseline confidence levels even after the correct answer is removed, with the probability mass assigned to distractors substantially exceeding that of the NOTA option. This reveals systematic overconfidence in which models commit to a distractor with nearly the same certainty as they would to the correct answer. Stratifying by baseline correctness (Appendix~\ref{appendix: statistical analysis}) reveals that models are significantly more likely to select NOTA on questions they answered correctly at baseline, suggesting that detection is linked to genuine knowledge rather than random guessing. However, the practical effect is marginal: even when models know the correct answer, they still fail to detect its absence from the candidate set.
% Most models achieve baseline accuracy well above 60\%, yet MCDR remains below 25\%, even worse than random chance. Further analysis (Appendix~\ref{sec:abstention_analysis}) shows that models default to the most plausible distractor, e.g., Qwen2.5-VL-7B 78\% of selections matching the highest-confidence incorrect option from the baseline. Meanwhile, high-confidence responses decrease under the multi-choice detection setting, and models are significantly more likely to select NOTA on questions answered correctly at baseline (OR: 2.7--5.5, $p < 10^{-10}$). This indicates a limited capacity to sense the absence of a correct option, but weak effect sizes ($\varphi = 0.13$--$0.26$) confirm that this signal rarely translates into reliable detection.

\noindent\textbf{Open-ended detection.} OEDR is consistently lower than MCDR across most models, indicating that models struggle even more when required to actively generate a rejection rather than select one from the candidates. Even when explicitly permitted to indicate that no option is correct, models overwhelmingly default to selecting from the provided candidates, possibly reflecting a forced-choice bias acquired from extensive training on multiple-choice tasks. A notable exception is Qwen2.5-Omni, which achieves an OEDR of 61.7\%; we discuss this outlier in Appendix~\ref{appendix: special mllms}.
% \noindent\textbf{Open-ended detection.} OEDR is consistently lower than MCDR across most models, indicating that models struggle even more when required to actively generate a rejection rather than select one from the candidates. Even when explicitly permitted to indicate that no option is correct, models overwhelmingly default to selecting from the provided candidates. This suggests that extensive training on multiple-choice tasks likely instills a strong forced-choice bias, in which models habitually commit to one of the given options without critically evaluating whether any candidate is truly correct. A notable exception is Qwen2.5-Omni, which achieves an OEDR of 61.7\%; we discuss this outlier in Appendix~\ref{appendix: special mllms}.

\noindent\textbf{Unprompted detection.} UDR is at or near 0\% across nearly all models, indicating that models inherently assume the correct answer must exist among the provided candidates and hardly spontaneously question the completeness of the option set. This result suggests that without explicit external cues, models lack the intrinsic capacity to recognize the absence of a valid answer.

% The three evaluation settings reveal that current MLLMs predominantly default to selecting the most plausible candidate rather than truly knowing the correct answer. The limited detection ability observed under the multi-choice and open-ended settings is largely contingent on explicit external cues, and vanishes almost entirely when models are left to their own judgment.

\subsection{Detection in Temporal Tasks}
\begin{table}[b]
\centering
\footnotesize
\setlength{\tabcolsep}{2pt}
\begin{tabular}{l cccc}
\toprule
Model & ACC & MCDR & OEDR & UDR \\
\midrule
Gemini & 74.6 / 60.2 & 29.1 / 40.4 & 34.6 / 52.1 & 3.6 / 0.0 \\
\midrule
Gemma3-4B & 60.0 / 33.9 & 0.0 / 2.3 & 1.8 / 0.6 & 0.0 / 0.0 \\
Gemma3-12B$^\dagger$ & 61.8 / 44.0 & 25.5 / 6.2 & 10.9 / 5.7 & 0.0 / 0.0 \\
Qwen2-VL & 69.1 / 45.2 & 3.6 / 1.1 & 1.8 / 0.0 & 0.0 / 0.0 \\
Qwen2.5-VL & 72.7 / 46.3 & 1.8 / 1.7 & 1.8 / 14.7 & 0.0 / 1.7 \\
Qwen2.5-Omni & 72.7 / 44.6 & 9.1 / 4.5 & 45.4 / 66.1 & 0.0 / 0.0 \\
Qwen3-VL & 80.0 / 55.4 & 9.1 / 2.3 & 7.3 / 4.0 & 0.0 / 0.0 \\
Qwen3-Omni & 74.6 / 50.6 & 5.5 / 5.7 & 14.6 / 18.6 & 1.8 / 0.6 \\
Mimo-VL & 78.2 / 47.5 & 30.9 / 33.3 & 1.8 / 2.8 & 0.0 / 0.0 \\
InternVL3 & 72.7 / 54.8 & 0.0 / 2.8 & 1.8 / 0.0 & 0.0 / 0.0 \\
InternVL3.5 & 76.4 / 48.0 & 3.6 / 3.4 & 1.8 / 0.6 & 0.0 / 0.6 \\
\bottomrule
\end{tabular}
\vspace{-5pt}
\caption{Absent answer detection on VideoMME temporal perception / temporal reasoning.}
\label{tab:videomme temporal}
\end{table}
Tables~\ref{tab:videomme temporal} report the detection results on the temporal perception and temporal reasoning subsets of VideoMME. 
On both temporal perception and temporal reasoning subsets, MCDR and OEDR drop substantially compared to the overall results in Table~\ref{tab:videomme} for most open-sourced models. This suggests that temporal tasks pose a greater challenge for absent answer detection.
%On temporal perception, most models achieve baseline accuracy above 70\%, yet MCDR and OEDR drop substantially compared to the overall results in Table~\ref{tab:videomme}, with several models falling below 5\% on both metrics. On temporal reasoning, baseline accuracy falls considerably, often below 50\%, and detection rates deteriorate even further. 

%We hypothesize that this degradation stems from the nature of distractors in temporal tasks: the incorrect options typically correspond to temporally adjacent events or causally related actions, making them inherently more plausible than distractors in non-temporal tasks. This elevated plausibility narrows the perceived gap between distractors and the absent ground-truth answer, leaving models more susceptible to committing to the best-matching candidate rather than recognizing that no valid option exists.

To examine whether this degradation is associated with more confusable distractors, we quantify the semantic similarity between each distractor and the removed ground-truth answer using two independent text embedding models, Qwen3-Embedding-0.6B~\cite{qwen3embedding} and bge-large-en-v1.5~\cite{bge_embedding}. For each instance, we compute both the mean similarity across distractors and the maximum distractor similarity. With Qwen3-Embedding-0.6B, the temporal subsets exhibit higher mean and maximum similarities (\(0.78\pm0.17\) and \(0.84\pm0.16\)) than the non-temporal subsets (\(0.66\pm0.13\) and \(0.73\pm0.13\)). The same pattern holds with bge-large-en-v1.5 (\(0.83\pm0.18/0.87\pm0.16\) vs. \(0.67\pm0.10/0.73\pm0.11\)). These results provide quantitative evidence that temporal questions contain semantically more confusable distractors.

\subsection{Impact of Frame Sampling Density}

\begin{figure}[t]
\centering
\includegraphics[width=1\linewidth]{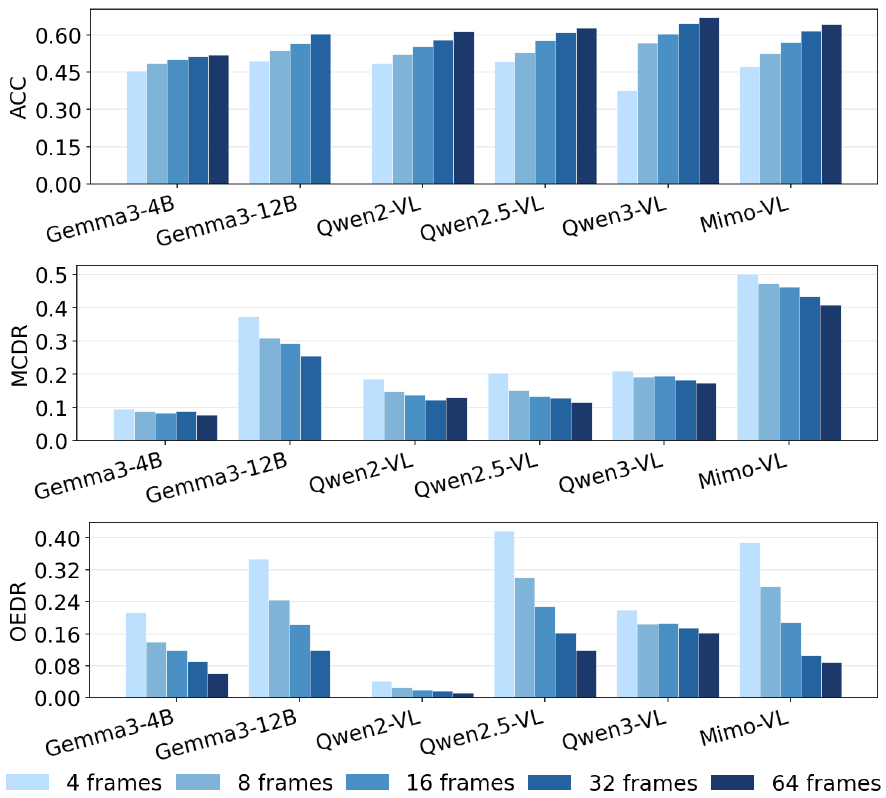}
\caption{ACC, MCDR, and OEDR on VideoMME across different frame sampling densities.}
\vspace{-20pt}
\label{fig: detection_vs_frames}
\end{figure}

As shown in Figure~\ref{fig: detection_vs_frames}, models achieve higher baseline accuracy with an increasing number of sampled frames. If the accuracy gains indicate that models have developed a more faithful understanding of the correct answer, denser sampling should likewise enable them to more reliably identify its absence from the candidate set. 
% However, our empirical results reveal the opposite trend. Both MCDR and OEDR consistently decrease with denser sampling, indicating that models become less likely to detect absent answers even as they better understand the video content. 
% This divergence reveals that denser visual input strengthens the model's ability to match candidates against video evidence, but simultaneously suppresses critical thinking rather than improving the model's capacity to detect the absent correct answer.

However, our empirical results reveal the opposite trend. Both MCDR and OEDR consistently decrease with denser sampling, indicating that models become less likely to detect absent answers even as baseline QA accuracy improves.

We further analyze the effect of increasing the number of input frames using Gemma3-12B. Repeating the last frame of an 8-frame input eight times yields MCDR/OEDR of (30.5/23.3), close to (30.9/24.5) with the original 8 frames. In contrast, using 16 distinct frames reduces the scores to (29.2/18.4). This comparison suggests that the degradation is more closely associated with the additional visual content than with the increased input length itself.
\section{CoT as a Mitigation Strategy}
\label{sec:cot}
The diagnostic findings above raise a natural question: can prompting strategies mitigate the absent answer detection failure? We investigate chain-of-thought (CoT) prompting as a potential intervention, instructing models to reason step-by-step about whether each candidate option is consistent with the video content before committing to a final answer. 
The detailed prompt can be found in Appendix~\ref{appendix: prompt}.
As shown in Table~\ref{tab:CoT}, CoT prompting yields substantial improvement for both InternVL3.5-8B-Instruct and Qwen3-VL-8B-Instruct, 
%with MCDR increasing from 17.4\% to 48.2\% (+30.8) and OEDR from 16.2\% to 49.9\% (+33.7), 
demonstrating that explicitly guiding models to evaluate each option against the video content can substantially recover detection capability. 
%However, detection rates remain below 50\% even with CoT, and the additional inference cost may limit practical applicability. These results suggest that CoT is a promising but insufficient mitigation strategy: while it partially unlocks latent detection capability, it does not compensate for the neglect of critical thinking ability during training.
As shown in Table~\ref{tab:CoT}, CoT prompting substantially improves absent-answer detection, with pronounced gains in both MCDR and OEDR across the two evaluated models. This suggests that explicitly encouraging option-wise verification can partially counteract the models' tendency to commit to plausible distractors. Nevertheless, the improvement remains insufficient: unprompted detection remains negligible, indicating that the ability to recognize an absent answer continues to depend heavily on explicit prompting. Moreover, these gains come at a substantial inference cost. For InternVL3.5, the average response length increases from 1.0/34.1 to 530.7/357.4 tokens under the multi-choice/open-ended settings, while for Qwen3-VL it increases from 7.7/23.6 to 1056.8/1112.5 tokens, respectively. CoT also slightly degrades baseline QA accuracy. Taken together, CoT constitutes an effective but incomplete inference-time mitigation: it improves prompted absent-answer detection, but neither resolves the underlying failure nor does so without compromising accuracy and generation efficiency.
% \begin{table}[h]
% \centering
% \footnotesize
% \setlength{\tabcolsep}{3pt}
% \begin{tabular}{l ccc}
% \toprule
% Model & MCDR & OEDR \\
% \midrule
% InternVL3.5 & 13.6 & 6.5  \\
% InternVL3.5 + CoT & \textbf{25.7} (\uparrow \textbf{12.1}) & \textbf{18.6} (\uparrow \textbf{12.1}) \\
% Qwen3-VL & 17.4 & 16.2\\
% Qwen3-VL + CoT & \textbf{48.2} (\uparrow \textbf{30.8}) & \textbf{49.9} (\uparrow \textbf{33.7}) \\
% \bottomrule
% \end{tabular}
% \vspace{-5pt}
% \caption{Effect of chain-of-thought prompting on absent answer detection on VideoMME. CoT substantially improves both MCDR and OEDR.}
% \vspace{-10pt}
% \label{tab: CoT}
% \end{table}
\begin{table}[h]
\centering
\footnotesize
\setlength{\tabcolsep}{3pt}
\begin{tabular}{l cccc}
\toprule
Model & ACC & MCDR & OEDR & UDR \\
\midrule
InternVL3.5
&\textbf{65.3}
& 13.6
& 6.5
& 0.0 \\

InternVL3.5 + CoT
& 63.1
& \textbf{25.7}
& \textbf{18.6}
& 0.0 \\
\midrule
Qwen3-VL
&\textbf{67.0}
& 17.4
& 16.2
& 0.7 \\

Qwen3-VL + CoT
& 63.9
& \textbf{48.2}
& \textbf{49.9}
& \textbf{4.4}\\
\bottomrule
\end{tabular}
\vspace{-5pt}
\caption{Effect of chain-of-thought prompting on absent answer detection on VideoMME. CoT substantially improves MCDR and OEDR, while slightly reducing ACC.}
\vspace{-10pt}
\label{tab:CoT}
\end{table}
% As shown in Table~\ref{tab: CoT}, CoT prompting yields a substantial improvement for Qwen3-VL, with MCDR increasing from 17.4\% to 48.2\% (+30.8) and OEDR from 16.2\% to 49.9\% (+33.7). This demonstrates that when explicitly guided to evaluate each option against the video content before selecting, models can substantially recover their detection capability. 
% However, several caveats temper the optimism of this result. First, even with CoT, detection rates remain below 50\%, meaning that models still fail to detect the absent answer in the majority of instances. 
% Third, CoT prompting introduces additional inference cost and latency, which may limit its practical applicability in real-time video understanding systems.
% These results suggest that CoT prompting is a promising but insufficient mitigation strategy. While it partially unlocks latent detection capability, it does not address the fundamental forced-choice bias instilled during training.

%%%%%%%%%%%%%%%%%%%%%%%%%%%%%%%
%%%%%%% conclusion
%%%%%%%%%%%%%%%%%%%%%%%%%%%%%%%
\section{Conclusion}
This work presents a systematic diagnostic study of absent answer detection for multimodal large language models in video understanding. By removing the ground-truth answer from the candidate set and evaluating models under three complementary settings,
%—multi-choice detection, open-ended detection, and unprompted detection—
we reveal a consistent and concerning pattern: current MLLMs overwhelmingly select plausible distractors rather than recognizing the absence of a valid answer. We further show that this failure is amplified in temporal reasoning tasks due to the heightened plausibility of semantically confusable distractors, and that increasing frame sampling density paradoxically suppresses detection rather than improving it. 
We also find that chain-of-thought prompting partially recovers detection capability, yet remains insufficient on its own, suggesting that addressing this limitation will require interventions beyond inference-time strategies.
These findings highlight a fundamental gap in the reliability of current video MLLMs and underscore the need for cultivating genuine video comprehension and critical thinking in future multimodal systems.

\newpage
\section*{Limitations}
Our study has several limitations. First, while we explore CoT prompting as a mitigation strategy, we do not investigate training-level interventions such as incorporating objectives aware of absent answers, which may more fundamentally address the forced-choice bias.
Second, while we cover a diverse set of models, the rapid pace of model development means that newer architectures may exhibit different behavior.

\section*{Acknowledgments}
This work was supported in part by NSF CNS-2112562 and
ARO W911NF-23-2-0224.

\bibliography{latex/custom}

@inproceedings{vllm_refuse,
  title={{Can Video LLMs Refuse to Answer? Alignment for Answerability in Video Large Language Models}},
  author={Yoon, Eunseop and Yoon, Hee Suk and Hasegawa-Johnson, Mark A and Yoo, Chang D},
  booktitle={Proceedings of the International Conference on Learning Representations (ICLR)},
  year={2025}
}

@inproceedings{nota_less_right,
    title = {{None of the Above, Less of the Right: Parallel Patterns in Human and {LLM} Performance on Multi-Choice Questions Answering}},
    author = {Tam, Zhi Rui  and
      Wu, Cheng-Kuang  and
      Lin, Chieh-Yen  and
      Chen, Yun-Nung},
    booktitle = "Findings of the Association for Computational Linguistics",
    year = "2025",
    pages = "20112--20134",
}

@inproceedings{UnsolvableProblemDetection,
  title={{Unsolvable Problem Detection: Robust Understanding Evaluation for Large Multimodal Models}},
  author={Miyai, Atsuyuki and Yang, Jingkang and Zhang, Jingyang and Ming, Yifei and Yu, Qing and Irie, Go and Li, Yixuan and Li, Hai Helen and Liu, Ziwei and Aizawa, Kiyoharu},
  booktitle={Proceedings of the 63rd Annual Meeting of the Association for Computational Linguistics (ACL)},
  pages={6497--6540},
  year={2025}
}

@article{gemini25,
  title={{Gemini 2.5: Pushing the Frontier with Advanced Reasoning, Multimodality, Long Context, and Next Generation Agentic Capabilities}},
  author={Gemini Team},
  journal={arXiv preprint arXiv:2507.06261},
  year={2025}
}

@article{gpt4o,
  title={{GPT-4o System Card}},
  author={OpenAI},
  journal={arXiv preprint arXiv:2410.21276},
  year={2024}
}

@article{gemma3,
      title={{Gemma 3 Technical Report}}, 
      author= {Gemma Team},
      year={2025},
      journal={arXiv preprint arXiv:2503.19786},
}

@article{qwen2vl,
  title={{Qwen2-VL: Enhancing Vision-Language Model's Perception of the World at Any Resolution}},
  author={Qwen Team},
  journal={arXiv preprint arXiv:2409.12191},
  year={2024}
}

@article{qwen25vl,
      title={{Qwen2.5-VL Technical Report}}, 
      author={Qwen Team},
      year={2025},
      journal={arXiv preprint arXiv:2502.13923},
}

@article{qwen3,
  title={{Qwen3 Technical Report}},
  author={Qwen Team},
  journal={arXiv preprint arXiv:2505.09388},
  year={2025}
}

@article{qwen3embedding,
  title={{Qwen3 Embedding: Advancing Text Embedding and Reranking through Foundation Models}},
  author={Qwen Team},
  journal={arXiv preprint arXiv:2506.05176},
  year={2025}
}

@article{bge_embedding,
      title={{C-Pack: Packaged Resources To Advance General Chinese Embedding}}, 
      author={Shitao Xiao and Zheng Liu and Peitian Zhang and Niklas Muennighoff},
      year={2023},
      journal={arXiv preprint arXiv:2309.07597},
}

@article{qwen3vl,
  title={{Qwen3-VL Technical Report}},
  author={Qwen Team},
  journal={arXiv preprint arXiv:2511.21631},
  year={2025}
}

@article{qwen25omni,
      title={{Qwen2.5-Omni Technical Report}}, 
      author={Qwen Team},
      year={2025},
      journal={arXiv preprint arXiv:2503.20215}
}

@article{qwen3omni,
  title={{Qwen3-Omni Technical Report}},
  author={Xu, Jin and Guo, Zhifang and Hu, Hangrui and Chu, Yunfei and Wang, Xiong and He, Jinzheng and Wang, Yuxuan and Shi, Xian and He, Ting and Zhu, Xinfa and others},
  journal={arXiv preprint arXiv:2509.17765},
  year={2025}
}

@article{qwen35omni,
  title={{Qwen3.5-Omni Technical Report}},
  author={Team, Qwen},
  journal={arXiv preprint arXiv:2604.15804},
  year={2026}
}

@article{mimo-vl,
      title={{MiMo-VL Technical Report}}, 
      author={Core Team and Zihao Yue and Zhenru Lin and Yifan Song and Weikun Wang and Shuhuai Ren and Shuhao Gu and Shicheng Li and Peidian Li and Liang Zhao and Lei Li and Kainan Bao and Hao Tian and Hailin Zhang and Gang Wang and Dawei Zhu and Cici and Chenhong He and Bowen Ye and Bowen Shen and Zihan Zhang and Zihan Jiang and Zhixian Zheng and Zhichao Song and Zhenbo Luo and Yue Yu and Yudong Wang and Yuanyuan Tian and Yu Tu and Yihan Yan and Yi Huang and Xu Wang and Xinzhe Xu and Xingchen Song and Xing Zhang and Xing Yong and Xin Zhang and Xiangwei Deng and Wenyu Yang and Wenhan Ma and Weiwei Lv and Weiji Zhuang and Wei Liu and Sirui Deng and Shuo Liu and Shimao Chen and Shihua Yu and Shaohui Liu and Shande Wang and Rui Ma and Qiantong Wang and Peng Wang and Nuo Chen and Menghang Zhu and Kangyang Zhou and Kang Zhou and Kai Fang and Jun Shi and Jinhao Dong and Jiebao Xiao and Jiaming Xu and Huaqiu Liu and Hongshen Xu and Heng Qu and Haochen Zhao and Hanglong Lv and Guoan Wang and Duo Zhang and Dong Zhang and Di Zhang and Chong Ma and Chang Liu and Can Cai and Bingquan Xia},
      year={2025},
      journal={arXiv preprint arXiv:2506.03569},
}

@article{internvl3,
  title={{InternVL3: Exploring Advanced Training and Test-Time Recipes for Open-Source Multimodal Models}},
  author={Zhu, Jinguo and Wang, Weiyun and Chen, Zhe and Liu, Zhaoyang and Ye, Shenglong and Gu, Lixin and Tian, Hao and Duan, Yuchen and Su, Weijie and Shao, Jie and others},
  journal={arXiv preprint arXiv:2504.10479},
  year={2025}
}

@article{internvl35,
  title={{InternVL3.5: Advancing Open-Source Multimodal Models in Versatility, Reasoning, and Efficiency}},
  author={Wang, Weiyun and Gao, Zhangwei and Gu, Lixin and Pu, Hengjun and Cui, Long and Wei, Xingguang and Liu, Zhaoyang and Jing, Linglin and Ye, Shenglong and Shao, Jie and others},
  journal={arXiv preprint arXiv:2508.18265},
  year={2025}
}

@inproceedings{videomme,
  title={{Video-MME: The First-Ever Comprehensive Evaluation Benchmark of Multi-Modal LLMs in Video Analysis}},
  author={Fu, Chaoyou and Dai, Yuhan and Luo, Yongdong and Li, Lei and Ren, Shuhuai and Zhang, Renrui and Wang, Zihan and Zhou, Chenyu and Shen, Yunhang and Zhang, Mengdan and others},
  booktitle={Proceedings of the IEEE/CVF Conference on Computer Vision and Pattern Recognition (CVPR)},
  pages={24108--24118},
  year={2025}
}

@article{longvideobench,
  title={{LongVideoBench: A Benchmark for Long-Context Interleaved Video-Language Understanding}},
  author={Wu, Haoning and Li, Dongxu and Chen, Bei and Li, Junnan},
  journal={Advances in Neural Information Processing Systems},
  volume={37},
  pages={28828--28857},
  year={2024}
}

@article{videommmu,
  title={{Video-MMMU: Evaluating Knowledge Acquisition from Multi-Discipline Professional Videos}},
  author={Hu, Kairui and Wu, Penghao and Pu, Fanyi and Xiao, Wang and Zhang, Yuanhan and Yue, Xiang and Li, Bo and Liu, Ziwei},
  journal={arXiv preprint arXiv:2501.13826},
  year={2025}
}

@inproceedings{mvbench,
  title={{MVBench: A Comprehensive Multi-Modal Video Understanding Benchmark}},
  author={Li, Kunchang and Wang, Yali and He, Yinan and Li, Yizhuo and Wang, Yi and Liu, Yi and Wang, Zun and Xu, Jilan and Chen, Guo and Luo, Ping and others},
  booktitle={Proceedings of the IEEE/CVF Conference on Computer Vision and Pattern Recognition (CVPR)},
  pages={22195--22206},
  year={2024}
}

@article{egoschema,
  title={{EgoSchema: A Diagnostic Benchmark for very Long-Form Video Language Understanding}},
  author={Mangalam, Karttikeya and Akshulakov, Raiymbek and Malik, Jitendra},
  journal={Advances in Neural Information Processing Systems},
  volume={36},
  pages={46212--46244},
  year={2023}
}

@article{mirage,
  title={{Mirage: The Illusion of Visual Understanding}},
  author={Asadi, Mohammad and O'Sullivan, Jack W and Cao, Fang and Nedaee, Tahoura and Rajabalifardi, Kamyar and Li, Fei-Fei and Adeli, Ehsan and Ashley, Euan},
  journal={arXiv preprint arXiv:2603.21687},
  year={2026}
}

@article{virtuebench,
  title={{VirtueBench: Evaluating Trustworthiness under Uncertainty in Long Video Understanding}},
  author={Yu, Xueqing and Li, Bohan and Li, Yan and Yang, Zhenheng},
  journal={arXiv preprint arXiv:2603.07071},
  year={2026}
}

@inproceedings{not_an_option,
  title={{Wait, that’s not an option: LLMs Robustness with Incorrect Multiple-Choice Options}},
  author={G{\'o}ral, Gracjan and Wi{\'s}nios, Emilia and Sankowski, Piotr and Budzianowski, Pawe{\l}},
  booktitle={Proceedings of the 63rd Annual Meeting of the Association for Computational Linguistics (ACL)},
  pages={1495--1515},
  year={2025}
}

@inproceedings{beyond_answers,
  title={{LLMs May Perform MCQA by Selecting the Least Incorrect Option}},
  author={Wang, Haochun and Zhao, Sendong and Qiang, Zewen and Xi, Nuwa and Qin, Bing and Liu, Ting},
  booktitle={Proceedings of the 31st International Conference on Computational Linguistics (COLING)},
  pages={5852--5862},
  year={2025}
}

@article{unkvqa,
  title={{UNK-VQA: A Dataset and a Probe into the Abstention Ability of Multi-Modal Large Models}},
  author={Guo, Yangyang and Jiao, Fangkai and Shen, Zhiqi and Nie, Liqiang and Kankanhalli, Mohan},
  journal={IEEE Transactions on Pattern Analysis and Machine Intelligence},
  volume={46},
  number={12},
  pages={10284--10296},
  year={2024},
  publisher={IEEE}
}

@inproceedings{zheng2024large,
  title={{Large Language Models are not Robust Multiple Choice Selectors}},
  author={Zheng, Chujie and Zhou, Hao and Meng, Fandong and Zhou, Jie and Huang, Minlie},
  booktitle={International Conference on Learning Representations},
  volume={2024},
  pages={19426--19454},
  year={2024}
}

@inproceedings{kvtp,
  title={{Keyframe-Oriented Vision Token Pruning: Enhancing Efficiency of Large Vision Language Models on Long-Form Video Processing}},
  author={Liu, Yudong and Sun, Jingwei and Lin, Yueqian and Zhang, Jianyi and Zhang, Jingyang and Yin, Ming and Wang, Qinsi and Li, Hai and Chen, Yiran},
  booktitle={2025 IEEE/CVF International Conference on Computer Vision (ICCV)},
  pages={20802--20811},
  year={2025},
  organization={IEEE}
}

@inproceedings{hippomm,
  title={{Hippomm: Hippocampal-Inspired Multimodal Memory for Long Audiovisual Event Understanding}},
  author={Lin, Yueqian and Zhang, Jingyang and Wang, Qinsi and Ye, Hancheng and Fu, Yuzhe and Liu, Yudong and Li, Hai Li and Chen, Yiran},
  booktitle={Proceedings of the IEEE/CVF Conference on Computer Vision and Pattern Recognition (CVPR)},
  pages={5968--5977},
  year={2026}
}

\appendix

%%%%%%%%%%%%%%%%%%%%%%%%%%%%%%%
%%%%%%% appendix
%%%%%%%%%%%%%%%%%%%%%%%%%%%%%%%
\newpage
\section{Appendix}

%%%%%%%%%%%%%%%%%%%%%%%%%%%%%%%
%%%%%%% prompt
%%%%%%%%%%%%%%%%%%%%%%%%%%%%%%%
\subsection{Prompt}
\label{appendix: prompt}
\label{sec:prompt appendix}
\begin{figure}[h]
\centering
\includegraphics[width=\linewidth]{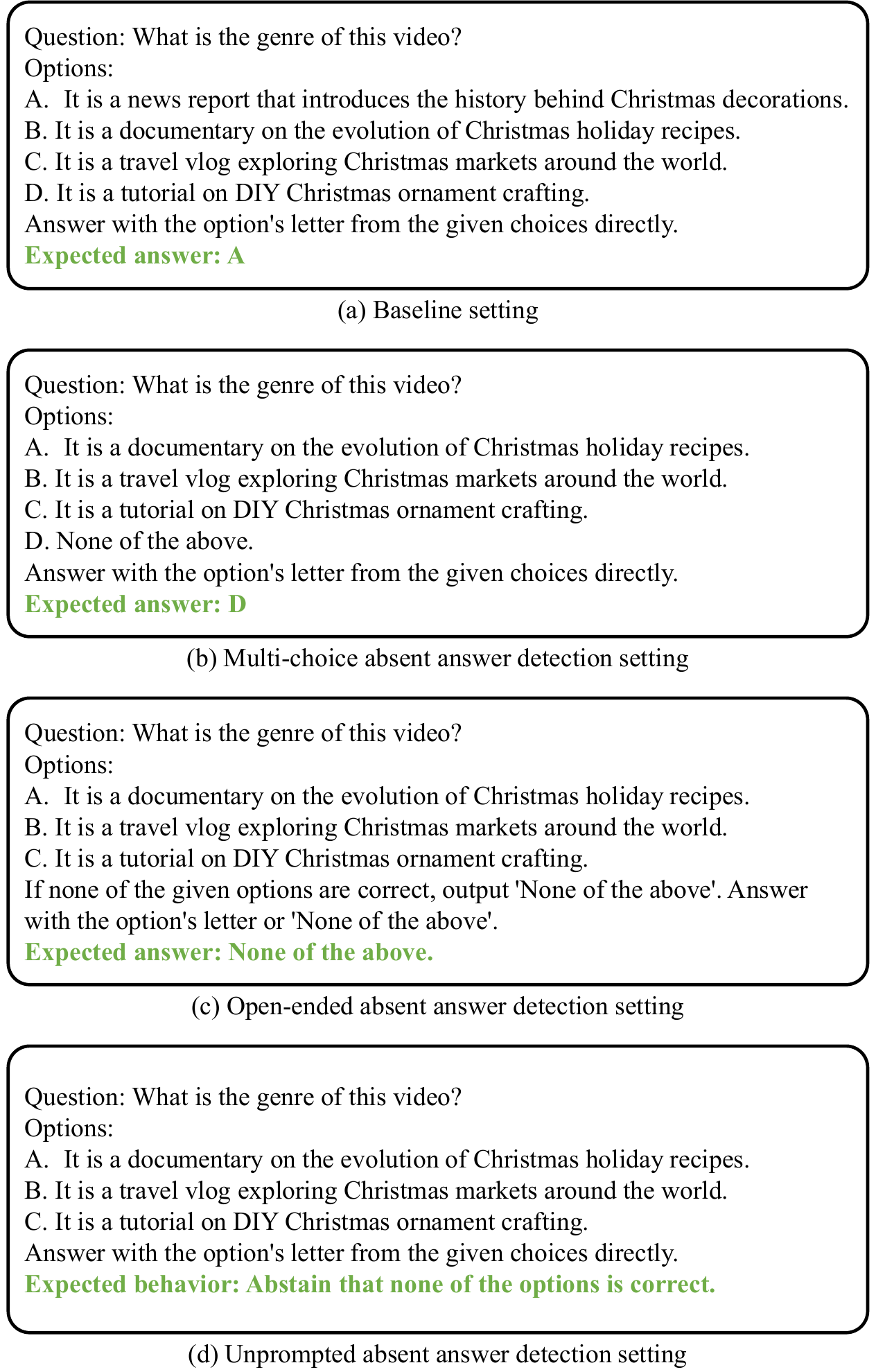}
\caption{Different prompt patterns for different experiment settings.}
\vspace{-10pt}
\label{fig:prompt}
\end{figure}

\begin{figure}[h]
\centering
\includegraphics[width=\linewidth]{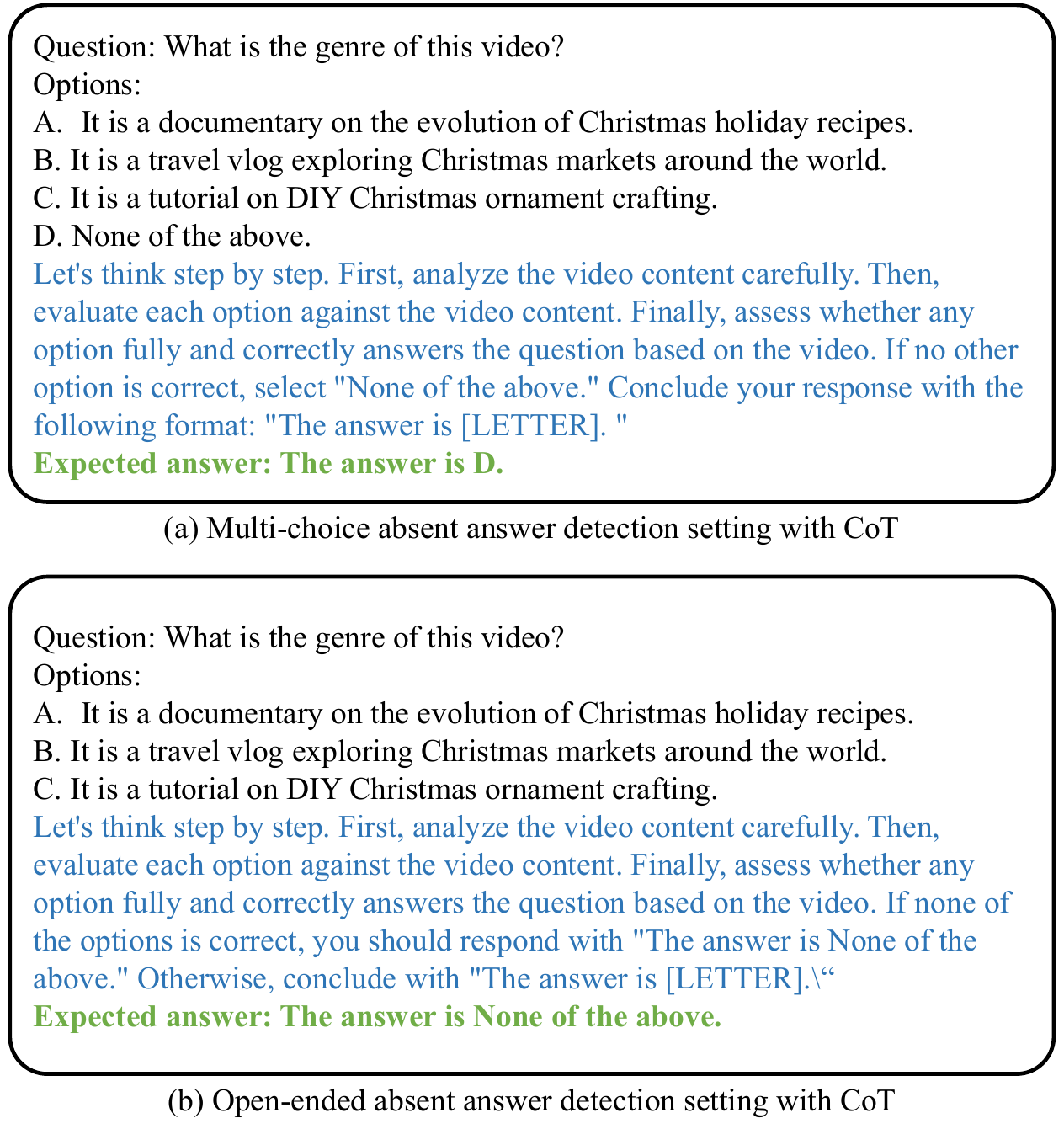}
\caption{Chain-of-thought prompt patterns for multi-choice and open-ended absent answer detection settings.}
\vspace{-10pt}
\label{fig:prompt_cot}
\end{figure}
Figure~\ref{fig:prompt} illustrates the prompt templates used across all four evaluation settings. In the \textbf{baseline setting (a)}, the model is presented with the original candidate set containing the ground-truth answer and asked to select an option directly. In the \textbf{multi-choice detection setting (b)}, the ground-truth answer is removed and a ``None of the above'' option is appended as an explicit detection cue. In the \textbf{open-ended detection setting (c)}, the ground-truth answer is removed and the instruction explicitly permits the model to output ``None of the above'' if it determines that no option is correct. In the \textbf{unprompted detection setting (d)}, the ground-truth answer is removed but no detection cue or additional instruction is provided. The model is simply asked to select from the remaining candidates under the standard protocol. Across all settings, the video input and question remain unchanged. Only the candidate set and instruction vary.

Figure~\ref{fig:prompt_cot} shows the CoT prompt templates used in Section~\ref{sec:cot}. In both the \textbf{multi-choice (a)} and \textbf{open-ended (b)} detection settings, the model is instructed to first analyze the video content, then evaluate each candidate option against it, and finally assess whether any option correctly answers the question. This step-by-step reasoning structure encourages the model to critically examine each option before committing to a selection, rather than directly pattern-matching to the most plausible candidate.

%%%%%%%%%%%%%%%%%%%%
%%%%% Position bias
%%%%%%%%%%%%%%%%%%%%
\subsection{Sensitivity to the Position of the NOTA Option}
\label{app:position}

Multiple-choice language models can be sensitive to option order.
To test whether the low MCDR is an artifact of the position of the
``None of the above'' (NOTA) option, we evaluate Qwen3-VL-8B while
placing NOTA at each position (A--D) in turn and randomly permuting
the remaining distractors.

\begin{table}[h]
\centering
\small
\begin{tabular}{lccccc}
\toprule
Model & A & B & C & D & Avg. \\
\midrule
Qwen3-VL-8B & 18.2 & 27.6 & 29.6 & 24.0 & 24.9 \\
\bottomrule
\end{tabular}
\caption{MCDR (\%) under different positions of the NOTA option.}
\end{table}

Although detection exhibits some position sensitivity, MCDR remains
low across all four placements (18.2--29.6\%). Thus, the overall
failure to detect absent answers is not an artifact of placing NOTA
at a particular option position.
%%%%%%%%%%%%%%%%%%%%%%%%%%%%%%%
%%%%%%% Confidence
%%%%%%%%%%%%%%%%%%%%%%%%%%%%%%%
\subsection{Confidence distribution}
\label{appendix: confidence distribution}
\begin{figure*}[htbp]
    \centering
\includegraphics[width=0.9\textwidth]{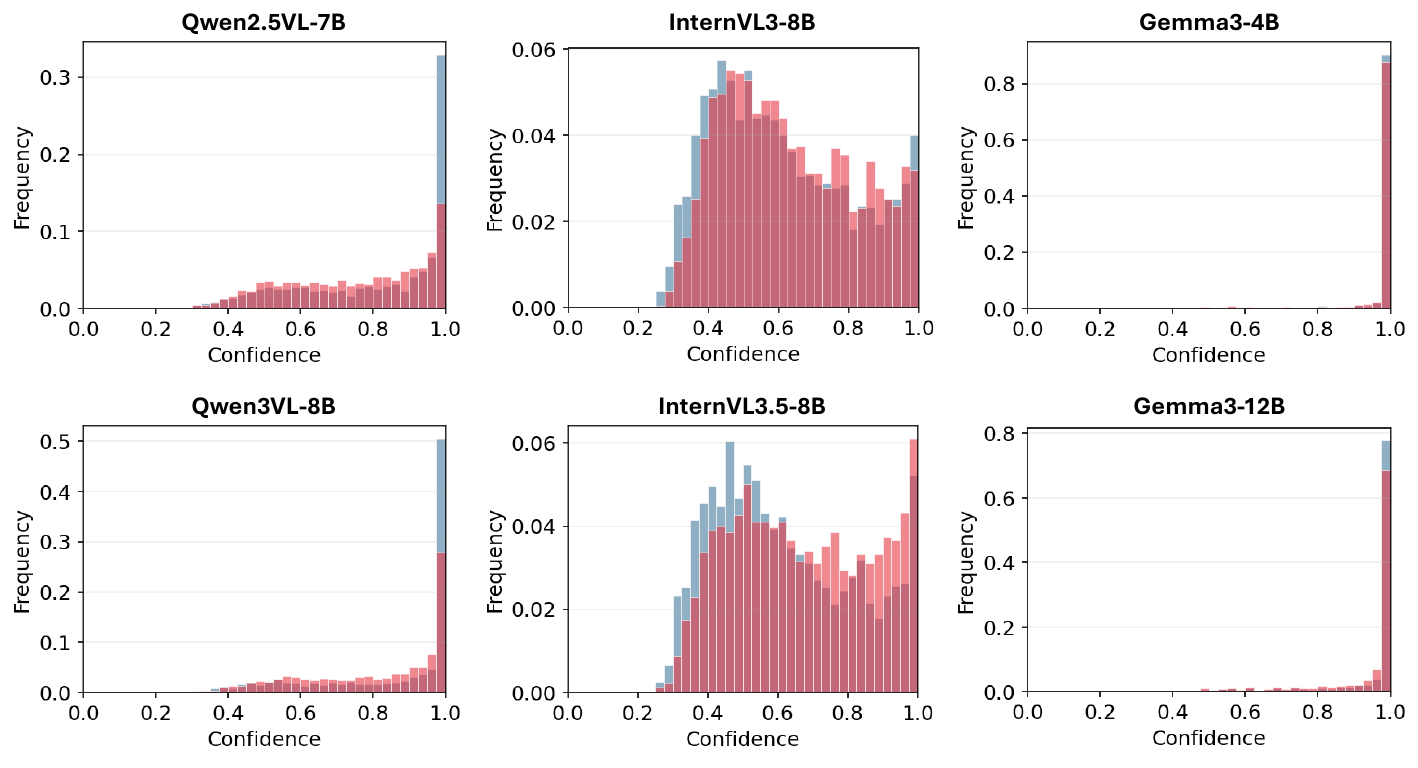}
    \vspace{-10pt}
    \caption{The confidence distribution under baseline setting and multi-choice detection setting.}
    \label{fig: baseline vs mc confidence}
    \vspace{-10pt}
\end{figure*}

\begin{figure*}[htbp]
    \centering
\includegraphics[width=0.9\textwidth]{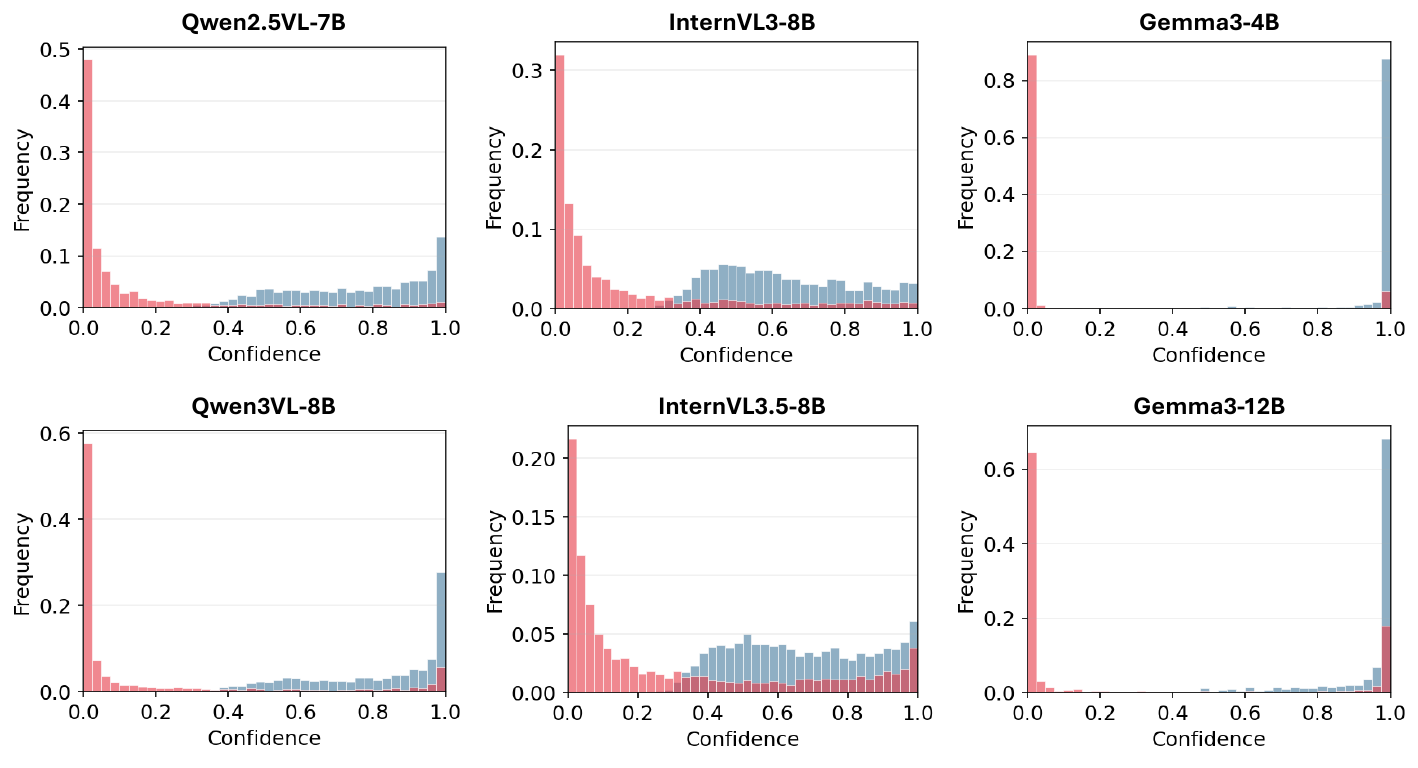}
    \vspace{-10pt}
    \caption{The confidence distribution of the "None of the above" option and the chosen option under multi-choice detection setting.}
    \vspace{-10pt}
    \label{fig: max vs none confidence}
\end{figure*}
To better understand why models fail to detect absent answers, we examine the confidence distributions under the baseline setting and the multi-choice detection setting. Confidence is defined as the probability assigned to the selected option, computed from the softmax-normalized logits over all candidate tokens.

Figure~\ref{fig: baseline vs mc confidence} compares the confidence distributions under the baseline setting (\textcolor{blue}{blue}) and the multi-choice detection setting (\textcolor{red}{red}). Across all six models, the two distributions exhibit substantial overlap, indicating that models maintain near-baseline confidence levels even after the ground-truth answer has been removed. This pattern is especially pronounced in Gemma3-4B and Gemma3-12B, where the vast majority of instances are concentrated near a confidence of 1.0 under both settings. In other words, removing the correct answer does not meaningfully reduce the model's confidence in its selected option. This reveals a systematic overconfidence in which models commit to a distractor with nearly the same certainty as they would commit to the correct answer.

Figure~\ref{fig: max vs none confidence} further decomposes the confidence under the multi-choice detection setting by comparing the probability assigned to the ``None of the above'' (NOTA) option (\textcolor{red}{red}) against the probability assigned to the chosen distractor (\textcolor{blue}{blue}). Across all models, the NOTA option receives overwhelmingly low confidence, with the majority of its probability mass concentrated near zero. In contrast, the chosen distractor consistently receives high confidence. This stark asymmetry suggests that models treat NOTA as a low-priority option rather than a genuine candidate for selection. Even when the correct answer is absent, models allocate minimal probability to the detection option and instead default to the most plausible distractor with high confidence. These findings confirm that current MLLMs lack the capacity to critically evaluate the completeness of the candidate set and instead exhibit a strong forced-choice bias toward selecting from the provided options.

%%%%%%%%%%%%%%%%%%%%%%%%%%%%%%%%
%%%%%%%% Statistical
%%%%%%%%%%%%%%%%%%%%%%%%%%%%%%%%
\subsection{Statistical Analysis}
\label{appendix: statistical analysis}
\begin{table*}[htbp]
\centering
\begin{tabular}{lcccc}
\toprule
Model & $p$-value ($\chi^2$) & $p$-value (Fisher) & Odds Ratio & $\phi$ \\
\midrule
Qwen2.5-VL   & $1.32 \times 10^{-23}$ & $2.32 \times 10^{-27}$ & 5.30 & 0.193 \\
Qwen2-VL     & $6.19 \times 10^{-15}$ & $4.63 \times 10^{-16}$ & 2.89 & 0.150 \\
Qwen3-VL     & $2.90 \times 10^{-31}$ & $1.79 \times 10^{-36}$ & 5.49 & 0.224 \\
Qwen2.5-Omni    & $6.58 \times 10^{-22}$ & $1.06 \times 10^{-23}$ & 3.06 & 0.185 \\
InternVL3.5  & $4.29 \times 10^{-17}$ & $4.96 \times 10^{-19}$ & 3.39 & 0.162 \\
InternVL3    & $3.55 \times 10^{-11}$ & $2.82 \times 10^{-12}$ & 2.71 & 0.127 \\
Mimo-VL      & $7.39 \times 10^{-43}$ & $1.17 \times 10^{-44}$ & 3.35 & 0.264 \\
Gemma3-4B       & $1.50 \times 10^{-18}$ & $4.41 \times 10^{-20}$ & 4.39 & 0.169 \\
Gemma3-12B      & $2.48 \times 10^{-42}$ & $6.16 \times 10^{-46}$ & 4.19 & 0.263 \\
\bottomrule
\end{tabular}
\caption{Statistical analysis of the association between selecting
``None of the above'' in the multi-choice detection setting and
answering correctly in the baseline setting.}
\label{tab:statistical_test}
\end{table*}

We investigate whether a model's ability to detect absent answers is related to its ability to answer the original question correctly. Specifically, we examine the association between two events for each instance: (1) the model selects the correct answer under the baseline setting, and (2) the model selects ``None of the above'' under the multi-choice detection setting. If models genuinely know the correct answer rather than guessing, they should be more likely to notice its absence when it is removed.

We conduct chi-squared tests and Fisher's exact tests of independence for each model on VideoMME. The results are reported in Table~\ref{tab:statistical_test}. All models yield highly significant $p$-values ($p < 10^{-10}$) under both tests, confirming that the two events are not independent. The odds ratios range from 2.7 to 5.5, indicating that models are substantially more likely to select the NOTA option when they answered the baseline question correctly. This suggests that models do possess a weak sensitivity to the absence of the correct answer, and that this sensitivity is linked to genuine knowledge rather than random guessing.

However, the effect sizes measured by Cramér's $\phi$ coefficient are consistently small, ranging from 0.13 to 0.26 across all models. This indicates that although the association is statistically significant, it explains only a small proportion of the variance in detection behavior. Even when a model knows the correct answer, it still overwhelmingly fails to detect its absence from the candidate set. The gap between statistical significance and practical effect size reinforces our main finding that current MLLMs lack reliable absent answer detection capability. Knowing the right answer is a necessary but far from sufficient condition for recognizing that it is missing.

\subsection{Analysis of Outlier Model Behaviors}
\label{appendix: special mllms}
% Two models exhibit notably divergent detection patterns that merit discussion.

% \noindent \textbf{MiMo-VL in multi-choice detection.} MiMo-VL achieves the highest MCDR across both benchmarks (40.9\% on VideoMME, 40.6\% on EgoSchema), substantially outperforming other open-sourced models. A distinguishing aspect of MiMo-VL's training is its Mixed On-policy Reinforcement Learning (MORL) framework~\citep{mimo-vl}, which incorporates Reinforcement Learning with Verifiable Rewards (RLVR) across multiple perception and reasoning tasks using rule-based reward functions. We hypothesize that this verification-oriented training encourages the model to critically assess the correctness of each candidate option rather than simply selecting the most plausible one, thereby making it more receptive to the NOTA option in the multi-choice detection setting. However, MiMo-VL's OEDR remains low (8.8\% on VideoMME), suggesting that this advantage is specific to the structured multi-choice format and does not generalize to open-ended detection.

\noindent\textbf{Qwen2.5-Omni in open-ended detection.} Qwen2.5-Omni achieves an anomalously high OEDR of 61.7\% on VideoMME, far exceeding all other models. However, closer inspection reveals that this result is accompanied by severe generation degeneration, casting doubt on whether the high detection rate reflects genuine absent answer detection. Across the evaluated instances, a substantial majority of the model's responses exhibit hallucinated multi-turn dialogue patterns: after producing an initial answer token, the model continues generating fabricated ``Human:'' turns that repeat the original question or degenerate into repetitive token loops. This behavior occurs regardless of whether the model selects ``None of the above'' or a distractor, suggesting that the model operates in an unstable decoding regime under this evaluation setting. Consequently, the elevated OEDR of Qwen2.5-Omni should be interpreted with caution, as the degenerate generation behavior undermines confidence that the model is genuinely detecting the absence of the correct answer through deliberate reasoning.

\noindent\textbf{MiMo-VL in multi-choice detection.} MiMo-VL exhibits a comparatively high MCDR relative to other open-source models. The released checkpoint is optimized with reinforcement learning using verifiable rewards, which may contribute to stronger reasoning or calibration under the forced-choice setting. However, without a matched pre-RL checkpoint or a controlled training ablation, we cannot isolate the source of this behavior. We therefore interpret the result as a model-specific outlier rather than evidence of a general preference for the multi-choice format.

\subsection{Implications for Future Work}
\label{appendix:discussion_solutions}

Our findings reveal a systematic gap between benchmark accuracy and genuine video understanding in current MLLMs. We discuss two directions that may help close this gap.

\paragraph{Benchmark design.} Current video understanding benchmarks predominantly assume that the correct answer is always present among the candidates, which risks overestimating model capabilities by rewarding plausible guessing. Future benchmarks should incorporate mechanisms that test whether models truly know the answer rather than simply selecting the most likely option. One approach is to include a proportion of questions where no correct option exists, requiring models to recognize the absence and abstain appropriately. Another approach is to increase the use of open-ended evaluation formats, where models must generate and justify their answers rather than selecting from a fixed set. Combining both answerable and unanswerable questions within the same benchmark would provide a more faithful assessment of model reliability, as high accuracy on answerable questions alone cannot distinguish genuine understanding from sophisticated pattern matching.

\paragraph{Training data.} Current training pipelines emphasize maximizing accuracy on standard question-answering tasks, which implicitly trains models to always commit to an answer regardless of the quality of the available options. To encourage more reliable behavior, training data should include examples that reward appropriate abstention. For instance, models should be exposed to instances where the provided options are all incorrect and the expected behavior is to refuse to select any of them. Models should also be trained on scenarios where the available information is insufficient to determine a definitive answer, learning to express uncertainty rather than fabricating a confident response.
%Such training would shift the optimization objective from always producing an answer to producing an answer only when the model has sufficient evidence, which better reflects genuine comprehension of video content.
Such training would shift the optimization objective from always producing an answer to producing an answer only when the model has sufficient evidence, which better reflects genuine comprehension of video content. Beyond data composition, future work could explicitly incorporate abstention-aware objectives that reward rejecting unsupported candidate sets and penalize confidently selecting an option when none is sufficiently supported. Developing such training-time interventions is a natural next step beyond the diagnostic scope of this work.
\end{document}